\documentclass{article}
\usepackage[margin=1in]{geometry}
\usepackage{microtype}
\usepackage{graphicx}
\usepackage{subfig}
\usepackage{booktabs} 
\usepackage{bbm}
\usepackage{listings}
\usepackage{algorithm}
\usepackage{algorithmic}
\usepackage{hyperref}
\usepackage{xcolor}

\newif\ifdraft \drafttrue

\newcommand{\cX}{\mathcal{X}}

\newcommand{\cA}{\mathcal{A}}
\newcommand{\ind}{\mathbbm{1}}

\usepackage{amssymb}
\usepackage{amsbsy}
\usepackage{amsmath}

\usepackage{amsthm}
\newtheorem{thm}{Theorem}[section] 
\theoremstyle{plain}

\newtheorem{theorem}[thm]{Theorem}
\theoremstyle{definition}
\newtheorem{lemma}[thm]{Lemma}
\newtheorem{definition}[thm]{Definition}

\usepackage{xr}
\usepackage{hyperref}
\usepackage{authblk}

\makeatletter
\newcommand*{\addFileDependency}[1]{
  \typeout{(#1)}
  \@addtofilelist{#1}
  \IfFileExists{#1}{}{\typeout{No file #1.}}
}
\makeatother

\title{Differentially Private Query Release Through Adaptive Projection}
\author[1]{Sergul Aydore}
\author[1,2]{William Brown}
\author[1,3]{Michael Kearns}
\author[1]{Krishnaram Kenthapadi}
\author[1]{Luca Melis}
\author[1,3]{Aaron Roth}
\author[1]{Ankit Siva}

\affil[1]{Amazon AWS AI/ML}
\affil[2]{Columbia University, New York, NY, USA}
\affil[3]{University of Pennsylvania, Philadelphia, PA, USA}

\begin{document}

\maketitle

\begin{abstract}
We propose, implement, and evaluate a new algorithm for releasing answers to very large numbers of statistical queries like $k$-way marginals, subject to differential privacy. Our algorithm makes adaptive use of a continuous relaxation of the \textit{Projection Mechanism}, which  answers queries on the private dataset using simple perturbation, and then attempts to find the synthetic dataset that most closely matches the noisy answers. We use a continuous relaxation of the synthetic dataset domain which makes the projection loss differentiable, and allows us to use efficient ML optimization techniques and tooling. Rather than answering all queries up front, we make judicious use of our privacy budget by iteratively finding queries for which our (relaxed) synthetic data has high error, and then repeating the projection. Randomized rounding allows us to obtain synthetic data in the original schema. 
We perform experimental evaluations across a range of parameters and datasets, and find that our method outperforms existing algorithms on large query classes.
\end{abstract}

\section{Introduction}
\label{sec:intro}
A basic problem in differential privacy is to accurately answer a large number $m$ of statistical queries (also known as \emph{linear} and \emph{counting} queries), which have the form, ``how many people in private dataset $D$ have property $P$?'' Marginal queries (also known as \emph{conjunctions}) --- which ask how many people in the dataset have particular combinations of feature values --- are one of the most useful and most studied special cases. The simplest technique for answering such queries is to compute each answer on the private dataset, and then perturb them with independent Gaussian noise. For a dataset of size $n$, this results in error scaling as $\tilde O\left(\frac{\sqrt{m}}{n}\right)$ \cite{delta}. This simple technique is useful for answering small numbers of queries. But it has been known since \cite{BLR08} that \emph{in principle}, it is possible to privately and accurately answer very large classes of queries (of size exponential in $n$), and that an attractive way of doing so is to encode the answers in a \emph{synthetic dataset}. Synthetic datasets have several advantages: most basically, they are a concise way of representing the answers to large numbers of queries. But they also permit one to evaluate queries \emph{other} than those that have been explicitly answered by the mechanism, and to take advantage of \emph{generalization}. Unfortunately, it is also known that improving on the error of the simple Gaussian perturbation technique is computationally hard in the worst case \cite{hardquery}. Moreover, constructing synthetic datasets is hard even when it would be possible to provide accurate answers with simple perturbation  \cite{hardsynthetic} for simple classes of queries such as the set of all $d \choose 2$ marginal queries restricted to 2 out of $d$ binary features (so-called $2$-way marginals).  As a result we cannot hope for a differentially private algorithm that can provably answer large numbers of statistical queries or generate interesting synthetic data in polynomial time. 

Nevertheless, there has been a resurgence of interest in private synthetic data generation and large-scale private queries due to the importance of the problem. Recent methods offer provable privacy guarantees, but have run-time and accuracy properties that must be evaluated empirically. 

\subsection{Our Contributions}

Our starting point is the (computationally inefficient) \emph{projection mechanism} of \cite{projection}, which is informally described as follows. We begin with a dataset $D \in \cX^n$. First, the values of each of the $m$ queries of interest $q_i$ are computed on the private dataset: $a = q(D) \in [0,1]^m$. Next, a privacy preserving vector of noisy answers $\hat a \in \mathbb{R}^m$ is computed using simple Gaussian perturbation. Finally, the vector of noisy answers $\hat a$ is \emph{projected} into the set of answer vectors that are consistent with some dataset to obtain a final vector of answers $a'$ --- i.e., the projection guarantees that $a' = q(D')$ for \emph{some} $D' \in \cX^n$. 
This corresponds to solving the optimization problem of finding the synthetic dataset $D' \in \cX^n$ that minimizes error $||q(D') - \hat a||_2$. This is known to be a near optimal mechanism for answering statistical queries \cite{projection}  but for most data and query classes, the projection step corresponds to a difficult discrete optimization problem. We remark that the main purpose of the projection is not (only) to construct a synthetic dataset, but to improve accuracy. 
This is analogous to how learning with a restricted model class like linear classifiers can improve accuracy if the data really is labeled by some linear function, i.e., the projection improves accuracy because by projecting into a set that contains the \emph{true} vector of answers $a$, it is imposing constraints that we know to be satisfied by the true (unknown) vector of answers.

Our core algorithm is based on a continuous relaxation of this projection step. This allows us to deploy first-order optimization methods, which empirically work very well despite the non-convexity of the problem. A further feature of this approach is that we can take advantage of sophisticated existing tooling for continuous optimization --- including autodifferentiation (to allow us to easily handle many different query classes) and GPU acceleration, which has been advanced by a decade of research in deep learning.
This is in contrast to related approaches like \cite{dualquery,Vietri2019NewOA} which use integer program solvers and often require designing custom integer programs for optimizing over each new class of queries.

We then extend our core algorithm by giving an adaptive variant that is able to make better use of its privacy budget, by taking advantage of generalization properties. 
Rather than answering \emph{all} of the queries up front, we start by answering a small number of queries, and then project them onto a vector of answers consistent with a relaxed synthetic dataset --- i.e., a dataset in a larger domain than the original data --- but one that still allows us to evaluate queries. At the next round, we use a private selection mechanism to find a small number of additional queries on which our current synthetic dataset performs poorly; we answer those queries, find a new synthetic dataset via our continuous projection, and then repeat. If the queries we have answered are highly accurate, then we are often able to find synthetic data representing the original data well after only having explicitly answered a very small number of them (i.e., we \emph{generalize} well to new queries). This forms a virtuous cycle, because if we only need to explicitly answer a very small number of queries, we can answer them highly accurately with our privacy budget. By taking our relaxed data domain to be the set of \emph{probability distributions} over one-hot encodings of the original data domain, we can finally apply randomized rounding to output a synthetic dataset in the original schema. 

We evaluate our algorithm on several datasets, comparing it to two state-of-the-art algorithms from the literature. A key advantage of our algorithm is that we can scale to large query workloads (in our experiments we answer roughly 20 million queries on some datasets and do not hit computational bottlenecks). We outperform the state of the art algorithm FEM (``Follow-the-Perturbed-Leader with Exponential Mechanism'') from \cite{Vietri2019NewOA}, which is one of the few previous techniques able to scale to large workloads. We also compare to algorithms that are unable to scale to large workloads, comparing to one of the state of the art methods, optimized variants of the HDMM (``High Dimensional Matrix Mechanism'') from \cite{mckenna2019graphical}. When run on a workload of roughly 65 thousand queries provided by the authors of \cite{mckenna2019graphical}, HDMM outperforms our algorithm. The result is an algorithm that we believe to be state of the art for large query workloads, albeit one that can  be outperformed for smaller workloads. 

\subsection{Additional Related Work}
Differential privacy offers a formal semantics for data privacy and was introduced by \cite{Dwork2006CalibratingNT}. The differential privacy literature is far too large to survey here; see \cite{privacybook} for a textbook introduction.

The problem of answering large numbers of queries on a private dataset (often via synthetic data generation) dates back to \cite{BLR08}. A line of early theoretical work \cite{BLR08,RR10,HR10,GRU12,projection} established statistical rates for answering very general classes of queries, showing that it is possible in principle (i.e., ignoring computation) to provide answers to \emph{exponentially many} queries in the size of the dataset.   This line of work establishes statistically optimal rates for the problem (i.e., matching statistical lower bounds), but provides algorithms that have running time that is generally \emph{exponential} in the data dimension, and hence impractical for even moderately high dimensional data. Moreover, this exponential running time is known to be necessary in the worst case \cite{DNRRV09,hardsynthetic,hardquery}. As a result, a line of work has emerged that tries to avoid this exponential running time in practice. The ``Multiplicative Weights Exponential Mechanism'' \cite{MWEM} uses optimizations to avoid exponentially large representations when the query class does not require it. Dwork, Nikolov, and Talwar give a theoretical analysis of a convex relaxation of the projection mechanism that can answer $k$-way marginals in time polynomial in $d^k$ --- albeit with accuracy that is sub-optimal by a factor of $d^{k/2}$ \cite{efficientmarginals}.``Dual Query'' \cite{dualquery} used a dual representation of the optimization problem implicitly solved by \cite{RR10,HR10,MWEM} to trade off the need to manipulate exponentially large state with the need to solve concisely defined but NP-hard integer programs. This was an ``oracle efficient'' algorithm.  The theory of oracle efficient synthetic data release was further developed in \cite{neel2019use}, and \cite{Vietri2019NewOA} give further improvements on oracle efficient  algorithms in this dual representation, and promising experimental results. We compare against the algorithm from \cite{Vietri2019NewOA} in our empirical results. We remark that marginal queries (the focus of our experimental evaluation) have been considered a canonical special case of the general query release problem, and the explicit focus of a long line of work \cite{marginals1,marginals2,marginals3,marginals4,marginals5}. 

A parallel line of work on \emph{matrix mechanisms} focused on optimizing error within a restricted class of mechanisms. Informally speaking, this class answers a specially chosen set of queries explicitly with simple perturbation, and then deduces the answers to other queries by taking linear combinations of those that were explicitly answered. One can optimize the error of this approach by optimizing over the set of queries that are explicitly answered \cite{matrix1}. Doing this exactly is also intractable, because it requires manipulating matrices that are exponential in the data dimension. This line of work too has seen heuristic optimizations, and the ``high dimensional matrix mechanism'' \cite{HDMM} together with further optimizations \cite{mckenna2019graphical} is able to scale to higher dimensional data and larger collections of queries --- although to date the size of the query classes that these algorithms can answer is smaller by several orders of magnitude compared to our algorithm and others in the oracle efficient line of work.

Finally, there is a line of work that has taken modern techniques for distribution learning (GANs, VAEs, etc.) and has made them differentially private, generally by training using private variants of stochastic gradient descent \cite{gan1,gan2,gan3,gan4,vae}. This line of work has shown some promise for image data as measured by visual fidelity, and for limited kinds of downstream machine learning tasks --- but generally has not shown promising results for enforcing consistency with simple classes of statistics like marginal queries. As a result we do not compare to approaches from this line of work.

\section{Preliminaries}
\label{sec:prelims}
\subsection{Statistical Queries and Synthetic Data}
Let $\cX$ be a data domain. In this paper, we will focus on data points containing $d$ categorical features: i.e. $\cX = \cX_1 \times \ldots \times \cX_d$, where each $\cX_i$ is a set of $t_i$ categories. A \emph{dataset} (which we will denote by $D$) consists of a multiset of $n$ points from $\cX$: $D \in \cX^n$. 

\begin{definition}[Statistical Query \cite{SQ}]
A \emph{statistical query} (also known as a \emph{linear query} or \emph{counting query}) is defined by a function $q_i:\cX \rightarrow [0,1]$. Given a dataset $D$, we will abuse notation and write $q_i(D)$ to denote the average value of the function $q_i$ on $D$:
$$q_i(D) = \frac{1}{n}\sum_{x \in D} q_i(x)$$
Given a collection of $m$ statistical queries $\{q_i\}_{i=1}^m$, we write $q(D) \in [0,1]^m$ to denote the vector of values $q(D) = (q_1(D),\ldots,q_m(D))$.
\end{definition}

An important type of statistical query is a $k$-way marginal, which counts the number of data points $x \in D$ that have a particular realization of feature values for some subset of $k$ features.\footnote{We define marginals for datasets with discrete features. In our experimental results we encode continuous features as discrete by standard binning techniques.}

\begin{definition}
A $k$-way marginal query is defined by a subset $S \subseteq [d]$ of $|S| = k$ features, together with a particular value for each of the features $y \in \prod_{i \in S} \cX_i$. Given such a pair $(S, y)$, let $\cX(S,y) = \{x \in \cX : x_i = y_i\ \ \forall i \in S\}$ denote the set of points that match the feature value $y_i$ for each of the $k$ features in $S$. The corresponding statistical query $q_{S,y}$ is defined as:
$$q_{S,y}(x) = \ind(x \in \cX(S,y))$$
Observe that for each collection of features (\emph{marginal}) $S$, there are $\prod_{i \in S}|\cX_i|$ many queries. 
\end{definition}

Given a set of $m$ statistical queries $q$, we will be interested in vectors of answers $a' \in [0,1]^m$ that represent their answers on $D$ \emph{accurately}:
\begin{definition}
Given a dataset $D$, a collection of $m$ statistical queries represented as $q:\cX^n\rightarrow [0,1]^m$, and a vector of estimated answers $a' \in [0,1]^m$, we say that $a'$ has $\ell_\infty$ or \emph{max} error $\alpha$ if $\max_{i \in [m]} |q_i(D)-a'_i| \leq \alpha$.
\end{definition}
In this paper we will represent vectors of estimated answers $a'$ \emph{implicitly} using some data structure $D'$ on which we can evaluate queries, and will write $q(D')$ for  $a'$. If $D' \in \cX^*$, then we refer to $D'$ as a \emph{synthetic dataset} --- but we will also make use of $D'$ lying in continuous relaxations of $\cX^n$ (and will define how query evaluation applies to such ``relaxed datasets'').

\subsection{Differential Privacy}
Two datasets $D, D' \in \cX^n$ are said to be \emph{neighboring} if they differ in at most one data point. We will be interested in \emph{randomized algorithms} $\cA:\cX^n\rightarrow R$ (where $R$ can be an arbitrary range).  
\begin{definition}[Differential Privacy \cite{Dwork2006CalibratingNT,delta}]
A randomized algorithm $\cA:\cX^n\rightarrow R$ is $(\epsilon,\delta)$ differentially private if for all pairs of neighboring datasets $D, D' \in \cX^n$ and for all measurable $S \subseteq R$:
$$\Pr[\cA(D) \in S] \leq \exp(\epsilon)\Pr[\cA(D') \in S] + \delta.$$ If $\delta = 0$ we say that $\cA$ is $\epsilon$-differentially private.
\end{definition}
Differential privacy is not convenient for tightly handling the degradation of parameters under composition, and so as a tool for our analysis, we use the related notion of (zero) Concentrated Differential Privacy:
\begin{definition}[Zero Concentrated Differential Privacy \cite{zCDP}]
An algorithm  $\cA:\cX^n\rightarrow R$ satisfies $\rho$-zero Concentrated Differential Privacy (zCDP) if for all pairs of neighboring datasets $D, D' \in \cX^n$, and for all $\alpha \in (0,\infty)$:
$$\mathbb{D}_\alpha(\cA(D),\cA(D')) \leq \rho \alpha$$
where $\mathbb{D}_\alpha(\cA(D),\cA(D'))$ denotes the $\alpha$-Renyi divergence between the distributions $\cA(D)$ and $\cA(D')$. 
\end{definition}
zCDP enjoys clean composition and postprocessing properties:
\begin{lemma}[Composition \cite{zCDP}]
\label{lem:composition}
Let $\cA_1:\cX^n\rightarrow R_1$ be $\rho_1$-zCDP. Let $\cA_2:\cX^n\times R_1 \rightarrow R_2$ be such that $\cA_2(\cdot, r)$ is $\rho_2$-zCDP for every $r \in R_1$. Then the algorithm $\cA(D)$ that computes $r_1 = \cA_1(D)$, $r_2 = \cA_2(D,r_1)$ and outputs $(r_1,r_2)$ satisfies $(\rho_1+\rho_2)$-zCDP.
\end{lemma}
\begin{lemma}[Post Processing \cite{zCDP}]
\label{lem:post}
Let $\cA:\cX^n \rightarrow R_1$ be $\rho$-zCDP, and let $f:R_1 \rightarrow R_2$ be an arbitrary randomized mapping. Then $f \circ \cA$ is also $\rho$-zCDP. 
\end{lemma}
Together, these lemmas mean that we can construct zCDP mechanisms by modularly combining zCDP sub-routines. Finally, we can relate differential privacy with zCDP:
\begin{lemma}[Conversions \cite{zCDP}]\
\label{lem:conversion}
\begin{enumerate}
\item If $\cA$ is $\epsilon$-differentially private, it satisfies $(\frac{1}{2}\epsilon^2)$-zCDP. 
\item If $\cA$ is $\rho$-zCDP, then for any $\delta > 0$, it satisfies $(\rho + 2\sqrt{\rho\log(1/\delta)},\delta)$-differential privacy.
\end{enumerate}
\end{lemma}

We will make use of two basic primitives from differential privacy, which we introduce here in the context of statistical queries. The first is the Gaussian mechanism.
\begin{definition}[Gaussian Mechanism]
The Gaussian mechanism $G(D,q_i,\rho)$ takes as input a dataset $D \in \cX^n$, a statistical query $q_i$, and a zCDP parameter $\rho$. It outputs $a_i = q_i(D) + Z$, where $Z \sim N(0,\sigma^2)$, where $N(0,\sigma^2)$ is the Gaussian distribution with mean $0$ and variance $\sigma^2 = \frac{1}{2 n^2 \rho}$. 
\end{definition}
\begin{lemma}[\cite{zCDP}]
\label{lem:gaussian}
For any statistical query $q_i$ and parameter $\rho > 0$, the Gaussian mechanism $G(\cdot,q_i,\rho)$ satisfies $\rho$-zCDP.
\end{lemma}

The second is a simple private ``selection'' mechanism called report noisy max --- we define a special case here, tailored to our use of it.
\begin{definition}[Report Noisy Max With Gumbel Noise]
\label{def:RNM}
The ``Report Noisy Max'' mechanism  $RNM(D,q,a,\rho)$ takes as input a dataset $D \in \cX^n$, a vector of $m$ statistical queries $q$, a vector of $m$ conjectured query answers $a$, and a zCDP parameter $\rho$. It outputs the index of the query with highest noisy error estimate. Specifically, it outputs $i^* = \arg\max_{i \in [m]}(|q_i(D) - a_i| + Z_i)$ where each $Z_i \sim \mathrm{Gumbel}\left(1/\sqrt{2\rho}n \right)$.
\end{definition}

\begin{lemma}
\label{lem:RNM}
For any vector of statistical queries $q$, vector of conjectured answers $a$, and zCDP parameter $\rho$, $RNM(\cdot,q,a,\rho)$ satisfies $\rho$-zCDP.
\end{lemma}
\begin{proof}
The report noisy max mechanism with Gumbel noise is equivalent to the exponential mechanism for sensitivity 1/n queries, and hence satisfies the bounded range property as defined in \cite{BR1}.  Lemma 3.2 of \cite{BR2} converts bounded range guarantees to zCDP guarantees, from which the claim follows. 
\end{proof}

\section{Relaxing the Projection Mechanism}
\label{sec:relax}
The projection mechanism of \cite{projection} can be described simply in our language. Given a collection of $m$ statistical queries $q$ and zCDP parameter $\rho$, it consists of two steps:
\begin{enumerate}
\item For each $i$, evaluate $q_i$ on $D$ using the Gaussian mechanism: $\hat a_i = G(D,q_i,\rho/m)$.
\item Find the synthetic dataset\footnote{In fact, in \cite{projection}, the projection is onto a set of datasets that allows datapoints to have positive or negative weights --- but their analysis also applies to projections onto the set of synthetic datasets in our sense. A statement of this can be found as Lemma 5.3 in \cite{proj2}.} $D'$ whose query values are closest to $\hat a$ in $\ell_2$ norm --- i.e., let $D' = \arg\min_{D' \in \cX^*} ||q(D') - \hat a||_2$.
\end{enumerate}
The output of the mechanism is the synthetic dataset $D'$, which implicitly encodes the answer vector $a' = q(D')$. Because the perturbation in Step 1 is Gaussian, and the projection is with respect to the $\ell_2$ norm, $D'$ is the maximum likelihood estimator for the dataset $D$ given the noisy statistics $\hat a$. The projection also serves to  enforce consistency constraints across all query answers, which perhaps counter-intuitively, is accuracy-improving. For intuition, the reader can consider the case in which all queries $q_i$ are identical: in this case, the scale of the initial Gaussian noise is $\Omega(\sqrt{m}/n)$, which is sub-optimal, because the single query of interest could have been privately answered with noise scaling only as $O(1/n)$. But the effect of the projection will be similar to \emph{averaging} all of the perturbed answers $\hat a_i$, because $q_i(D')$ will be constrained to take a fixed value across all $i$ (since the queries are identical), and the mean of a vector of noisy estimates minimizes the Euclidean distance to those estimates. This has the effect of averaging out much of the noise, recovering error $O(1/n)$.  The projection mechanism is easily seen to be $\rho$-zCDP --- the $m$ applications of $(\rho/m)$-zCDP instantiations of the Gaussian mechanism in Step 1 compose to satisfy $\rho$-zCDP by the composition guarantee of zCDP (Lemma \ref{lem:composition}), and Step 2 is a post-processing operation, and so by Lemma \ref{lem:post} does not increase the privacy cost. This mechanism is nearly optimal amongst the class of all differentially private mechanisms, as measured by $\ell_2$ error, in the worst case over the choice of statistical queries \cite{projection}. Unfortunately, Step 2 is in general an intractable computation, since it is a minimization of a non-convex and non-differentiable objective over an exponentially large discrete space. The first idea that goes into our algorithm (Algorithm \ref{alg:proj}) is to relax the space of datasets $\cX^n$ to be a continuous space, and to generalize the statistical queries $q_i$ to be differentiable over this space. Doing so allows us to apply powerful GPU-accelerated tools for differentiable optimization to the projection step 2.

\paragraph{From Categorical to Real Valued Features}
Our first step is to embed categorical features into \emph{binary} features using a one-hot encoding. This corresponds to replacing each categorical feature $\cX_i$ with $t_i$ binary features  $\cX_i^1\times\ldots\times\cX_i^{t_i} = \{0,1\}^{t_i}$, for each $x \in \cX$. Exactly one of these new $t_i$ binary features corresponding to categorical feature $i$ is set to 1 for any particular data point $x \in \cX$: If $x_i = v_j$ for some $v_j \in \cX_i$, then we set $\cX_i^j = 1$ and $\cX_i^{j'} = 0$ for all $j' \neq j$. Let $d' = \sum_{i=1}^d t_i$ be the dimension of a feature vector that has been encoded using this one-hot encoding. Under this encoding, the datapoints $x$ are embedded in the binary feature space $\{0,1\}^{d'}$. We will aim to construct synthetic data that lies in a continuous relaxation of this binary feature space. For example, choosing $\cX^{r} = [0,1]^{d'}$ is natural. In our experiments, we choose $\cX^{r} =[-1,1]^{d'}$, which empirically leads to an easier optimization problem. We further apply a SparseMax~\cite{martins2016softmax} transformation to convert this relaxed data domain into the set of (sparse) probability distributions over one-hot encodings. In addition to improving accuracy, this transformation allows us to apply randomized rounding to recover a dataset in the original schema: we discuss this further in Section~\ref{sec:sparse_max_projection}.

Let $h:\cX\rightarrow \{0,1\}^{d'}$ represent the function that maps a $x \in \cX$ to its one-hot encoding. We abuse notation and for a dataset $D \in \cX^n$, write $h(D)$ to denote the one-hot encoding of every $x \in D$.

\paragraph{From Discrete to Differentiable Queries}
Consider a marginal query $q_{S,y}:\cX\rightarrow \{0,1\}$ defined by some $S \subseteq [d]$ and $y \in \prod_{i\in S} \cX_i$. Such a query can be evaluated on a vector of categorical features $x \in \cX$ in our original domain. Our goal is to construct an  \emph{equivalent extended differentiable query} $\hat q_{S,y}:\cX^r\rightarrow \mathbb{R}$ that has two properties:
\begin{definition}[Equivalent Extended Differentiable Query]
Given a statistical query $q_i : \cX\rightarrow [0,1]$, we say that $\hat q_i : \cX^r\rightarrow \mathbb{R}$ is an extended differentiable query that is equivalent to $q_i$ if it satisfies the following two properties:
\begin{enumerate}
    \item $\hat q_i$ is differentiable over $\cX^r$ --- i.e. for every $x \in \cX^r$, $\nabla q_i(x)$ is defined, and
    \item $\hat q_i$ agrees with $q_i$  on every feature vector that results from a one-hot encoding. In other words, for every $x \in \cX$: $q_i(x) = \hat q_{i} (h(x))$. 
\end{enumerate}
\end{definition}
We will want to give equivalent extended differentiable queries for the class of $k$-way marginal queries. Towards this end, we define a product query:
\begin{definition}
Given a subset of features $T \subseteq [d']$, the product query $q_T:\cX^r\rightarrow \mathbb{R}$ is defined as:
$q_T(x) = \prod_{i \in T} x_i$.
\end{definition}
By construction, product queries satisfy the first  requirement for being extended differentiable queries: they are defined over the entire relaxed feature space $\cX^r$, and are differentiable (since they are monomials over a real valued vector space). It remains to observe that for every marginal query $q_{S,y}$, there is an equivalent product query $\hat q_{S,y}$ that takes value $q_{S,y}(x)$ on the one-hot encoding $h(x)$ of $x$ for every $x$. 
\begin{lemma}
Every $k$-way marginal query has an equivalent extended differentiable query in the class of product queries. In other words, for every $k$-way marginal query $q_{S,y}:\cX^n\rightarrow \{0,1\}$, there is a corresponding product query $\hat q_{S,y} = q_T(y):\cX^r\rightarrow \mathbb{R}$ with $|T| = k$ such that for every $x \in \cX$: $q_{S,y}(x) = q_T(h(x))$. 
\label{test}
\end{lemma}
\begin{proof}
We construct $T$ in the straightforward way: for every $i \in S$, we include in $T$ the coordinate corresponding to $y_i \in \cX_i$. Now consider any $x$ such that $q_{S,y}(x) = 1$. It must be that for every $i \in S$, $x_i = y_i$. By construction, the product $q_T(h(x)) = \prod_{j \in T} h(x)_j = 1$ because all terms in the product evaluate to 1. Similarly, if $q_{S,y}(x) = 0$, then it must be that for at least one coordinate $j \in T$, $h(x)_j = 0$, and so $q_T(h(x)) = \prod_{j \in T} h(x)_j = 0$.
\end{proof}

\section{The Relaxed Adaptive Projection (RAP) Mechanism}
We here introduce the ``Relaxed Adaptive Projection'' (RAP) mechanism (Algorithm \ref{alg:main}), which has three hyper-parameters: the \emph{number of adaptive rounds} $T$, the \emph{number of queries per round} $K$, and the \emph{size of the (relaxed) synthetic dataset} $n'$. In the simplest case, when $T = 1$ and $K = m$, we recover the natural relaxation of the projection mechanism:
\begin{enumerate}
    \item We evaluate each query $q_i \in Q$ on $D$ using the Gaussian mechanism to obtain a noisy answer $\hat a_i$, and
    \item Find a \emph{relaxed} synthetic dataset $D' \in X^r$ whose equivalent extended differentiable query values are closest to $\hat a$ in $\ell_2$ norm: $D' = \arg\min_{D' \in (\cX^r)^{n'}}||\hat q(D') - \hat a||_2$.
\end{enumerate}
Because step 2 is now optimizing a continuous, differentiable function over a continuous space (of dimension $d' \cdot n'$, we can use existing tool kits for performing the optimization -- for example, we can use auto-differentiation tools, and optimizers like Adam~\cite{kingma2015adam}. (Recall that the projection is a post-processing of the Gaussian mechanism, and so the privacy properties of the algorithm are independent of our choice of optimizer). Here $n'$ is a hyperparameter that we can choose to trade off the expressivity of the synthetic data with the running-time of the optimization: If we choose $n' = n$, then we are assured that it is possible to express $D$ exactly in our relaxed domain: as we choose smaller values of $n'$, we introduce a source of representation error, but decrease the dimensionality of the optimization problem in our projection step, and hence improve the run-time of the algorithm. In this simple case, we can recover an accuracy theorem by leveraging the results of \cite{projection}:
\begin{theorem}
Fix privacy parameters $\epsilon,\delta > 0$,  a synthetic dataset size $n'$, and any set of $m$ $k$-way product queries $q$. 
If the minimization in the projection step is solved exactly, then the average error for the RAP mechanism when $T = 1$ and $K = m$ can be bounded as:
$$\sqrt{\frac{1}{m}||q(D) - q(D')||_2^2} \leq$$$$ O\left(\frac{(d'(\log k + \log n')+\log(1/\beta)\ln(1/\delta))^{1/4}}{\sqrt{\epsilon n}} +  \frac{\sqrt{\log k}}{\sqrt{n'}} \right)$$
with probability $1-\beta$ over the realization of the Gaussian noise.
\label{thm:accuracy_theorem}
\end{theorem}
See Appendix \ref{sec:proof_accuracy_theorem} for proof.

This is an ``oracle efficient'' accuracy theorem in the style of \cite{dualquery,Vietri2019NewOA,oraclenonconvex} in the sense that it assumes that our heuristic optimization succeeds (note that this assumption is not needed for the privacy of our algorithm, which we establish in Theorem \ref{thm:privacy}). Compared to the accuracy theorem for the FEM algorithm proven in \cite{Vietri2019NewOA}, our theorem improves by a factor of $\sqrt{d'}$. 

In the general case, our algorithm runs in $T$ rounds: After each round $t$, we have answered some \emph{subset} of the queries $Q_S \subseteq Q$, and perform a projection only with respect to the queries in $Q_S$ for which we have estimates, obtaining an intermediate relaxed synthetic dataset $D_t'$. At the next round, we augment $Q_S$ with $K$ additional queries $q_i$ from $Q \setminus Q_S$ chosen (using report noisy max) to maximize the disparity $|q_i(D_t') - q_i(D)|$. We then repeat the projection. In total, this algorithm only explicitly answers $T\cdot K$ queries, which might be $\ll m$. But by selectively answering queries for which the consistency constraints imposed by the projection with respect to previous queries have not correctly fixed, we aim to expend our privacy budget more wisely. Adaptively answering a small number of ``hard'' queries has its roots in a long theoretical line of work \cite{RR10,HR10,GRU12}. 

\begin{algorithm}[h]
\caption{Relaxed Projection (RP)}
\label{alg:proj}
\begin{algorithmic}
    \STATE {\bfseries Input:} A vector of differentiable queries $q:\cX^r\rightarrow \mathbb{R}^{m'}$, a vector of target answers $\hat a \in \mathbb{R}^{m'}$, and an initial dataset $D' \in (\cX^r)^{n'}$.
    \STATE Use any differentiable optimization technique (Stochastic Gradient Descent, Adam, etc.) to attempt to find:
    $$D_S = \arg\min_{D' \in (\cX^r)^{n'}} ||q(D') - \hat a||_2^2$$
    \STATE Output $D_S$.
\end{algorithmic}
\end{algorithm}

\begin{algorithm}[h]
    \caption{Relaxed Adaptive Projection (RAP)}
    \label{alg:main}
\begin{algorithmic}
    \STATE {\bfseries Input:} A dataset $D$, a collection of $m$ statistical queries $Q$, a ``queries per round'' parameter $K \leq m$,  a ``number of iterations'' parameter $T \leq m/K$, a synthetic dataset size $n'$, and differential privacy parameters $\epsilon,\delta$.
    \STATE Let $\rho$ be such that:
    $$\epsilon = \rho + 2\sqrt{\rho\log(1/\delta)}$$
    \IF{$T = 1$}
      \FOR{$i = 1$ to $m$}
        \STATE Let $\hat a_i = G(D,q_i,\rho/m)$.
      \ENDFOR
      \STATE Randomly initialize $D' \in (\cX^r)^{n'}$.
      \STATE Output $D'= RP(q, \hat a, D')$.
      \ELSE
      \STATE Let $Q_S = \emptyset$ and $D'_0 \in (\cX^r)^{n'}$ be an arbitrary initialization.
      \FOR{$t = 1$ to $T$}
          \FOR{$k = 1$ to $K$}
            \STATE Define $\hat q^{Q \setminus Q_S}(x) = (\hat q_i(x) : q_i \in Q \setminus Q_S)$ where $\hat q_i$ is an equivalent extended differentiable query for $q_i$.
             \STATE Let $q_i = RNM(D,\hat q^{Q \setminus Q_S},\hat q^{Q \setminus Q_S}(D'_{t-1}),\frac{\rho}{2T\cdot K})$.
             \STATE Let $Q_S = Q_S \cup \{q_i\}$.
             \STATE Let $\hat a_i = G(D,q_i,\frac{\rho}{2T\cdot K})$.
             \ENDFOR
             \STATE Define $q^{Q_S}(x) = (q_i(x) : q_i \in Q_S)$ and $\hat a = \{\hat a_i : q_i \in Q_S\}$ where $\hat q_i$ is an equivalent extended differentiable query for $q_i$.
            Let $D'_t = RP(q^{Q_S}, \hat a, D_{t-1}')$. 
      \ENDFOR
      \STATE Output $D'_T$.
    \ENDIF
\end{algorithmic}
\end{algorithm}

\begin{theorem}\label{thm:privacy} For any query class $Q$, any set of parameters $K,T,n'$, and any privacy parameters $\epsilon,\delta > 0$, the
RAP mechanism $RAP(\cdot,Q,K,T,n',\epsilon,\delta)$ (Algorithm \ref{alg:main}) is $(\epsilon,\delta)$-differentially private.
\end{theorem}
See Appendix \ref{sec:proof_privacy} for proof.

\paragraph{Randomized Rounding to Output a Synthetic Dataset}
\label{sec:sparse_max_projection}

We use the SparseMax~\cite{martins2016softmax} transformation to generate relaxed synthetic data in which each set of one-hot columns, corresponding to the original features, is normalized to a (sparse) probability distribution.
More specifically, after each step of the optimization technique in Algorithm.~\ref{alg:proj}, we apply SparseMax independently to each set of encoded columns in the synthetic dataset $D_{S}$.
Randomized rounding (i.e. for each feature independently, selecting a one-hot encoding with probability proportional to its probability in the relaxed synthetic data) can then be applied to produce a synthetic dataset consistent with the original schema. 
This preserves the expected value of marginal queries, and can preserve their values exactly in the limit as we take multiple samples. As we show in our experiments, preserving the worst case error over many marginals requires only moderate oversampling in practice (5 samples per data point).

\section{Empirical Evaluation}

\subsection{Implementation and Hyperparameters}
We implement\footnote{\href{https://www.github.com/amazon-research/relaxed-adaptive-projection}{github.com/amazon-research/relaxed-adaptive-projection}} Algorithm~\ref{alg:main} in Python \cite{python}, using the JAX library~\cite{jax2018github} for auto-differentiation of queries and the Adam optimizer~\cite{kingma2015adam} (with learning rate $0.001$) for the call to RP (Algorithm~\ref{alg:proj}).
For each call to RP, we do early stopping if the relative improvement on the loss function between consecutive Adam steps is less than $10^{-7}$.
The number of maximum Adam steps per RP round is set to $5000$.
Fig.~\ref{fig:python_snippet} contains a Jax code snippet, which computes 3-way product queries on a dataset $D$. A benefit of using JAX (or other packages with autodifferentiation capabilities) is that to instantiate the algorithm for a new query class, all that is required is to write a new python function which computes queries in the class --- we do not need to perform any other reasoning about the class. In contrast, approaches like \cite{dualquery,Vietri2019NewOA} require deriving an integer program to \emph{optimize} over each new class of interest, and approaches like \cite{mckenna2019graphical} require performing an expensive optimization over each new workload of interest. This makes our method more easily extensible. 

\begin{figure}
\begin{lstlisting}[language=Python, linewidth=\columnwidth,breaklines=true, basicstyle=\small]
import jax.numpy as np
def threeway_marginals(D):
    return np.einsum('ij,ik,il->jkl', D, D, D)/D.shape[0]

\end{lstlisting}
\caption{Python function used to compute 3-way product queries}
\label{fig:python_snippet}
\end{figure}
JAX also has the advantages of being open source and able to take advantage of GPU acceleration. We run our experiments for Algorithm~\ref{alg:main} on an EC2 p2.xlarge instance (1 GPU, 4 CPUs, 61 GB RAM). 
    For FEM we use the code from the authors of~\cite{Vietri2019NewOA} available at~\url{https://github.com/giusevtr/fem}, using the hyperparameters given in their tables 2 and 3 for the experimental results we report in Figures \ref{fig:epsilons} and \ref{fig:workloads}, respectively. Their code requires the Gurobi integer program solver; we were able to obtain a license to Gurobi for a personal computer, but not for  EC2 instances, and so we run FEM on  a 2019 16" MacBook Pro (6 CPUs, 16GB RAM) (Gurobi does not support GPU acceleration) --- as a result we do not report timing comparisons. We remark that an advantage of our approach is that it can leverage the robust open-source tooling (like JAX and Adam) that has been developed for deep learning, to allow us to easily take advantage of large-scale distributed GPU accelerated  computation.  

For HDMM+LSS and HDMM+PGM implementations, we used code provided by the authors of \cite{mckenna2019graphical} which was hard-coded with a query strategy for a  particular set of 62876 marginal queries on the Adult dataset, which we also run on a MacBook Pro.

For most experiments, we set the size of the synthetic data $n' = 1000$ --- significantly smaller than $n$ for both of our datasets (see Table \ref{tab:datasets}). See Appendix \ref{sec:extraplots} for an investigation of performance as a function of $n'$. For the remaining hyperparameters $K$ and $T$, 
we optimize over a small grid of values  (see Table~\ref{tab:hyperparams}) and report the combination with the smallest error. This is also how error is reported for FEM. For all experiments we optimize over $\cX^r = [-1,1]^{d'}$, which empirically had better convergence rates compared to using $\cX^r = [0,1]^{d'}$ --- likely because gradients of our queries vanish at $0$. 

\subsection{Selecting Marginals}
For our main set of experiments comparing to the FEM algorithm of \cite{Vietri2019NewOA}, we mirror their experimental design in \cite{Vietri2019NewOA}, and given $k$, we select a number of marginals $S$ (i.e., subsets of categorical features), referred to as the {\it workload}, at random, and then enumerate all queries consistent with the selected marginals (i.e., we enumerate all $y \in \prod_{i \in S} \cX_i$). For each experiment, we fix the query selection process and random seed so that both algorithms in our comparisons are evaluated on exactly the same set of queries.
See Fig.~\ref{fig:queries} in Appendix \ref{sec:extraplots} for the total number of selected queries across different workloads on both of our datasets, which vary in a range between $10^5$ and $10^8$. For our comparison to the HDMM variants of \cite{mckenna2019graphical}, we compare on the particular set of 62876 3-way marginal queries on Adult for which the hard-coded query strategy in their provided code is optimized on.

\subsection{Experimental Results}
We evaluate both our algorithm and FEM on the two datasets used by \cite{Vietri2019NewOA} in their evaluation: ADULT and LOANS ~\cite{Dua2019}.

Just as in~\cite{Vietri2019NewOA}, both datasets are transformed so that all features are categorical --- real valued features are first bucketed into a finite number of categories. The algorithms are then run on a one-hot encoding of the discrete features, as we described in Section \ref{sec:relax}. To ensure consistency, we use the pre-processed data exactly as it appears in their repository for \cite{Vietri2019NewOA}. See Table~\ref{tab:datasets} for a summary of the datasets. 
\begin{table}
    \centering
    {\begin{tabular}{cccc}
    Dataset & Records & Features & Transformed Binary Features\\
     \hline
     ADULT &  48842 &  15 & 588 \\
     LOANS &  42535 &  48 & 4427 \\
    \end{tabular}}
\caption{Datasets. Each dataset starts with the given number of original (categorical and real valued) features. After our transformation, it is encoded as a dataset with a larger number of binary features.}
\label{tab:datasets}
\end{table}

\begin{table}
    \centering
    {\begin{tabular}{ccc}
    Parameter & Description & Values \\
    \hline
    $K$ & Queries per round & 5 10 25 50 100\\
    $T$ & Number of iterations & 2 5 10 25 50\\
    \end{tabular}}
\caption{RAP hyperparameters tested in our experiments}
\label{tab:hyperparams}
\end{table}

We mirror the evaluation in \cite{Vietri2019NewOA} and focus our experiments comparing to FEM on answering 3-way and 5-way marginals. 

We also compare to the  High Dimensional Matrix Mechanism (HDMM) with Local Least Squares (HDMM+LLS) and Probabilistic Graphical Model (HDMM+PGM) inference from \cite{mckenna2019graphical}, but these mechanisms do not scale to large workloads, and the existing implementations are hard-coded with optimizations for a fixed set of queries on Adult. Hence in our comparison to HDMM+LSS and HDMM+PGM, we can only run these algorithms on the fixed set of 62876 3-way marginals defined on the Adult dataset that the code supports. 

We use the maximum error between answers to queries on the synthetic data and the correct answers on the real data across queries ($\max_i | q_i(D’) - q_i(D) |$) as a performance  measure. For calibration, we also report a naive baseline corresponding to the error obtained by answering every query with ``0''. 
Error above this naive baseline is uninteresting. 
For all experiments, we fix the privacy parameter $\delta$ to $\frac{1}{n^2}$, where $n$ is the number of records in the dataset, and vary $\epsilon$ as reported.

\begin{figure}[t]
        \center
        \subfloat[][\small{ADULT dataset on 3-way marginals}]{\includegraphics[width=.488\columnwidth,trim={0.5cm 0.5cm 0.5cm 0.5cm}]{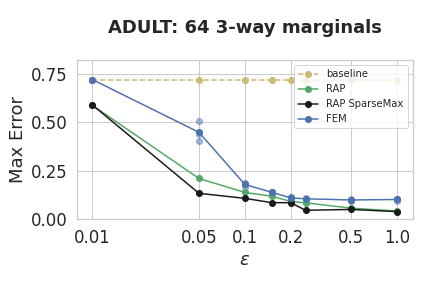}}  \hfill 
        \subfloat[][\small{LOANS dataset on 3-way marginals} ]{\includegraphics[width=.488\columnwidth,trim={0.5cm 0.5cm 0.5cm 0.5cm}]{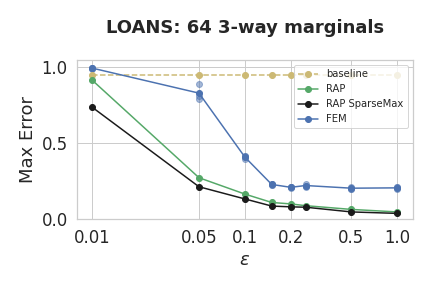}} \\
        \subfloat[][\small{ADULT dataset on 5-way marginals} ]{\includegraphics[width=.488\columnwidth]{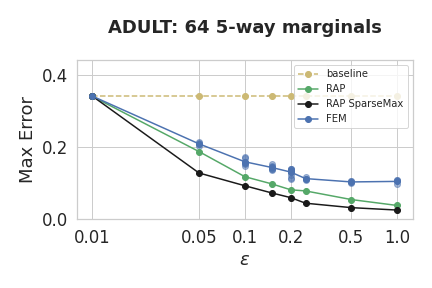}} \hfill
        \subfloat[][\small{LOANS dataset on 5-way marginals} ]{\includegraphics[width=.488\columnwidth]{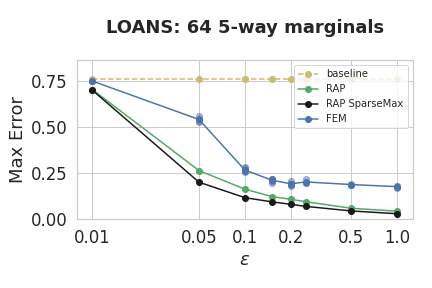}}
        
        \caption{Max-error for 3 and 5-way marginal queries on different privacy levels.  The number of marginals is fixed at 64.}
    \label{fig:epsilons}
\end{figure}

\begin{figure}[h!]
        \center
        \subfloat[][\small{ADULT dataset on 3-way marginals}]{\includegraphics[width=.488\columnwidth,trim={0.5cm 0.5cm 0.5cm 0.5cm}]{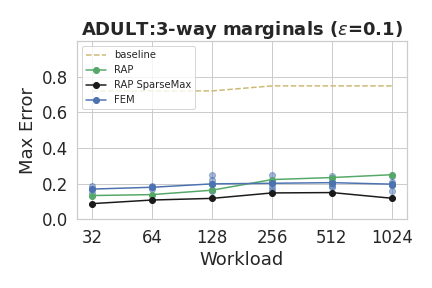}}  \hfill 
        \subfloat[][\small{LOANS dataset on 3-way marginals} ]{\includegraphics[width=.488\columnwidth,trim={0.5cm 0.5cm 0.5cm 0.5cm}]{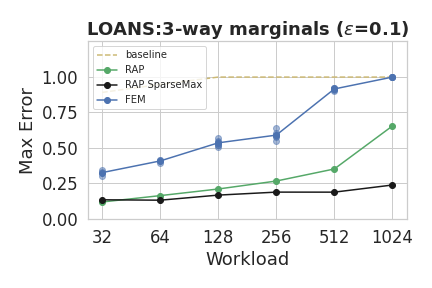}} \\
        \subfloat[][\small{ADULT dataset on 5-way marginals} ]{\includegraphics[width=.488\columnwidth]{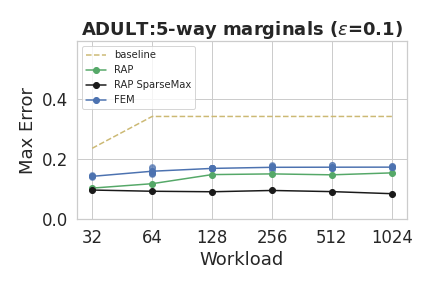}} \hfill
        \subfloat[][\small{LOANS dataset on 5-way marginals} ]{\includegraphics[width=.488\columnwidth]{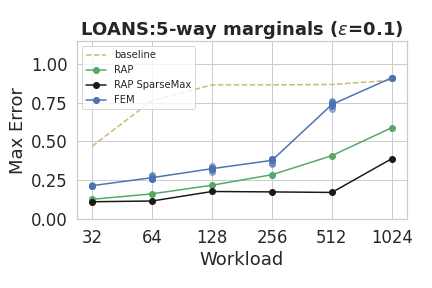}} 

        \caption{Max error for increasing number of 3 and 5-way marginal queries with $\epsilon = 0.1$}
    \label{fig:workloads}
\end{figure}

\begin{figure}[h!]
        \center
        \subfloat[][\small{ADULT dataset on 3-way marginals}]{\includegraphics[width=.488\columnwidth,trim={0.5cm 0.5cm 0.5cm 0.5cm}]{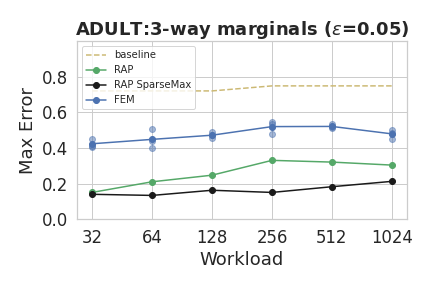}}  \hfill 
        \subfloat[][\small{LOANS dataset on 3-way marginals} ]{\includegraphics[width=.488\columnwidth,trim={0.5cm 0.5cm 0.5cm 0.5cm}]{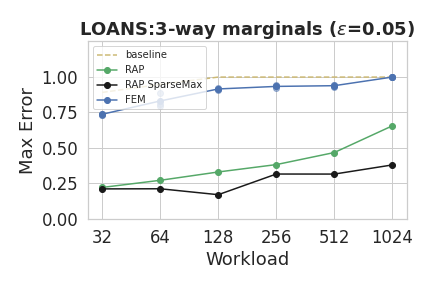}} \\
        \subfloat[][\small{ADULT dataset on 5-way marginals} ]{\includegraphics[width=.488\columnwidth]{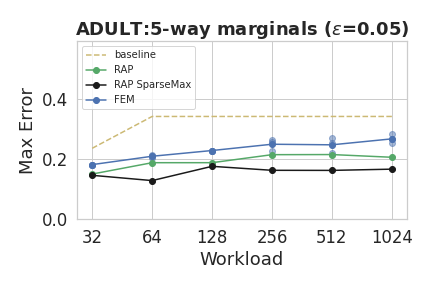}} \hfill
        \subfloat[][\small{LOANS dataset on 5-way marginals} ]{\includegraphics[width=.488\columnwidth]{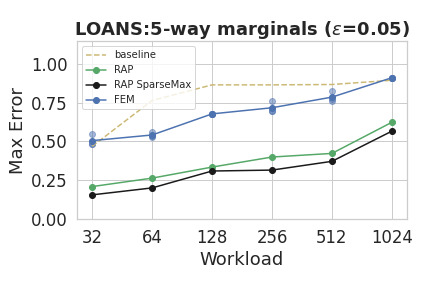}} 

        \caption{Max error for increasing number of 3 and 5-way marginal queries with $\epsilon=0.05$}
    \label{fig:workloads0_05}
\end{figure}

\begin{figure}[h!]
        \center
        \subfloat[][\small{}]{\includegraphics[width=.488\columnwidth,trim={0.5cm 0.5cm 0.5cm 0.5cm}]{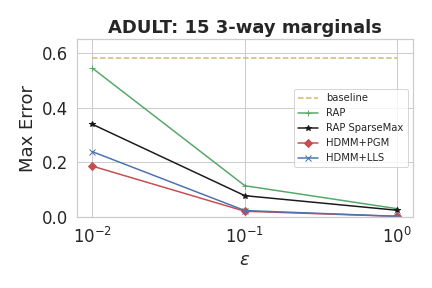}}  \hfill 
        \subfloat[][\small{} ]{\includegraphics[width=.488\columnwidth,trim={0.5cm 0.5cm 0.5cm 0.5cm}]{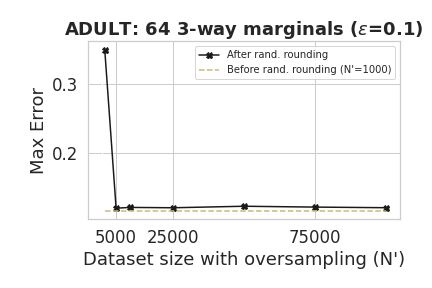}} 
    \caption{(a):Max-error of HDMM variants and RAP for the set of 15 3-way marginal queries on ADULT provided by  \cite{mckenna2019graphical} at  different privacy levels. (b) Max Error of RAP before and after randomized rounding with different levels of oversampling.}
    \label{fig:amalgam}
\end{figure}

In Figs.~\ref{fig:epsilons}-\ref{fig:amalgam}(a) we show how our performance scales with the privacy budget $\epsilon$ for a fixed number of marginals. Figs.~\ref{fig:workloads},~\ref{fig:workloads0_05} show our performance for a fixed
privacy budget as we increase the number of marginals being preserved. 

We significantly outperform FEM in all comparisons considered, and performance is particularly strong in the important high-privacy and high workload regimes (i.e., when $\epsilon$ is small and $m$ is large).
However, both HDMM+PGM and HDMM+LLS outperform RAP in the small workload regime in the comparison we are able to run. 

Figure~\ref{fig:amalgam} (b) shows how randomized rounding, when applied on the synthetic dataset generated by RAP and SparseMax, affects the error on the marginals for different levels of oversampling.
The error after randomly rounding each data point $5$ times  (obtaining a synthetic dataset of size $n' = 5,000$) approaches the error before applying randomized rounding and slowly converges for larger oversampling rates.

\begin{figure}[t]
\center
        \subfloat[][]{\includegraphics[width=.5\columnwidth,trim={0.5cm 0.5cm 0.5cm 0.5cm}]{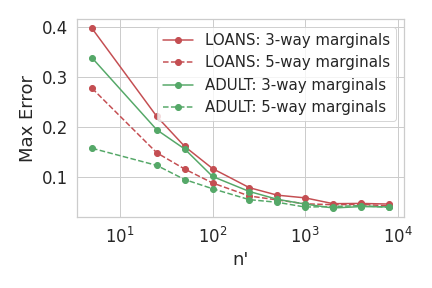}}    \subfloat[][]{\includegraphics[width=.5\columnwidth,trim={0.5cm 0.5cm 0.5cm 0.5cm}]{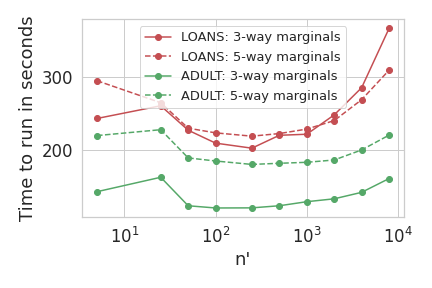}} 
    \caption{(a) Error and (b) run-time as a function of the synthetic dataset size $n'$.}
    \label{fig:runtimesmall}
\end{figure}

We also investigate the run-time and accuracy of our algorithm as a function of the synthetic dataset size $n'$ --- see Figure \ref{fig:runtimesmall}, and Appendix \ref{sec:extraplots} for more details. Here we note two things: (i) We can take $n'$ quite small as a function of the true dataset size $n$, until a certain point (below $n' = 1000$) at which point error starts increasing, (ii) Run time also decreases with $n'$, until we take $n'$ quite small, at which point the optimization problem appears to become more difficult. 

Finally, as we have noted already, an advantage of our approach is its easy extensibility: to operate on a new query class, it is sufficient to write the code to evaluate queries in that class. To demonstrate this, in the Appendix we plot results for a different query class: linear threshold functions. 

\section{Conclusion}
We have presented a new, extensible method for privately answering large numbers of statistical queries, and producing synthetic data consistent with those queries. Our method relies on a continuous, differentiable relaxation of the projection mechanism, which allows us to use existing powerful tooling developed for deep learning. We demonstrate on a series of experiments that our method  out-performs existing techniques across a wide range of parameters in the large workload regime.

\clearpage
\bibliography{ref}
\bibliographystyle{plain}

\appendix
\section{Proof of Theorem \ref{thm:accuracy_theorem}}
\label{sec:proof_accuracy_theorem}
\begin{proof}
We reduce to the (unrelaxed) projection mechanism, which has the following guarantee proven by \cite{projection}: for any dataset $D$ consisting of $n$ elements from a \emph{finite} data universe $\cX$, and for any set of $m$ statistical queries $q$, the projection mechanism results in a dataset $D'$ such that: $\sqrt{\frac{1}{m}||q(D')-q(D)||_2^2} \leq \alpha$ for $$\alpha = O\left(\frac{(\ln(|\cX|/\beta)\ln(1/\delta))^{1/4}}{\sqrt{\epsilon n}}\right).$$
Consider a  finite data universe $\cX^\eta = \{0, \eta, 2\eta, \ldots, 1\}^{d'}$ for some discretization parameter $0 < \eta < 1/k$. Given a dataset $D' \in \cX^r$, let $D'_\eta \in \cX^{\eta}$ be the dataset that results from ``snapping'' each real-valued $x \in D$ to its closest discrete valued point $x_\eta \in \cX^r$. Observe that by construction, $||x-x(\eta)||_\infty \leq \eta$, and as a result, for $k$-way product query $q_i$, we have $|q_i(D') - q_i(D'_\eta)| \leq O(\eta k)$. Now let $\hat D' = \arg\min_{\hat D' \in (\cX^r)^*}||a - q(\hat D')||$ and $D'' = \arg\min_{D'' \in (\cX^\eta)^*}||a - q(D'')||$. From above, we know that $\sqrt{\frac{1}{m}||q(D'') - q(\hat D')||} \leq O(\eta k)$, and hence from an application of the triangle inequality, we have that $\sqrt{\frac{1}{m}||q(D) - q(\hat D')||_2^2} \leq O\left(\frac{(\ln(|\cX^\eta|/\beta)\ln(1/\delta))^{1/4}}{\sqrt{\epsilon n}} + \eta k\right)$. Finally, for any dataset $\hat D' \in (\cX^r)^*$, there exists a dataset $D' \in (\cX^r)^{n'}$ such that $\sqrt{\frac{1}{m}||q(D') - q(\hat D')||_2^2} \leq O(\frac{\sqrt{\log k}}{\sqrt{n'}})$ (This follows from a sampling argument, and is proven formally in  \cite{BLR08}.) Hence, a final application of the triangle inequality yields:
$$\sqrt{\frac{1}{m}||q(D) - q(D')||_2^2} \leq$$$$  O\left(\frac{(\ln(|\cX^\eta|/\beta)\ln(1/\delta))^{1/4}}{\sqrt{\epsilon n}} + \eta k + \frac{\sqrt{\log k}}{\sqrt{n'}} \right)$$

Choosing $\eta = \frac{\sqrt{\log k}}{k \sqrt{n'}}$ and noting that $|\cX^\eta| = (\frac{1}{\eta})^{d'}$ yields the bound in our theorem.

\end{proof}

\section{Proof of Theorem \ref{thm:privacy}}
\label{sec:proof_privacy}
\begin{proof}
The privacy of Algorithm \ref{alg:main} follows straightforwardly from the tools we introduced in Section \ref{sec:prelims}. First consider the case of $T = 1$. The algorithm makes $m$ calls to the Gaussian mechanism, each of each satisfies $\rho/m$-zCDP by construction and Lemma \ref{lem:gaussian}. In combination, this satisfies $\rho$-zCDP by the composition Lemma (Lemma \ref{lem:composition}). It then makes a call to the relaxed projection algorithm $RP$, which is a postprocessing of the Gaussian mechanism, and hence does not increase the zCDP parameter, by Lemma \ref{lem:post}. Hence the algorithm is $\rho$-zCDP, and by our choice of $\rho$ and Lemma \ref{lem:conversion}, satisfies $(\epsilon,\delta)$ differential privacy. 

Now consider the case of $T > 1$. Each iteration of the inner loop makes one call to report noisy max, and one call to the Gaussian mechanism. By construction and by Lemmas \ref{lem:gaussian} and \ref{lem:RNM}, each of these calls satisfies $\frac{\rho}{2TK}$-zCDP, and together by the composition Lemma \ref{lem:composition}, satisfy $\frac{\rho}{TK}$-zCDP. The algorithm then makes a call to the relaxed projection algorithm $RP$, which is a post-processing of the composition of the Gaussian mechanism with report noisy max, and so does not increase the zCDP parameter by Lemma \ref{lem:post}. The inner loop runs $T\cdot K$ times, and so the entire algorithm satisfies $\rho$-zCDP by the composition Lemma \ref{lem:composition}. By our choice of $\rho$ and Lemma \ref{lem:conversion}, our algorithm satisfies $(\epsilon,\delta)$ differential privacy as desired. 
\end{proof}

\section{Additional Plots}
\label{sec:extraplots}
\begin{figure}[t]
    \begin{tabular}{cc}
    \includegraphics[width=.48\columnwidth]{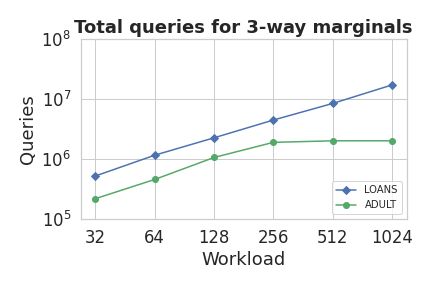} & \includegraphics[width=.48\columnwidth]{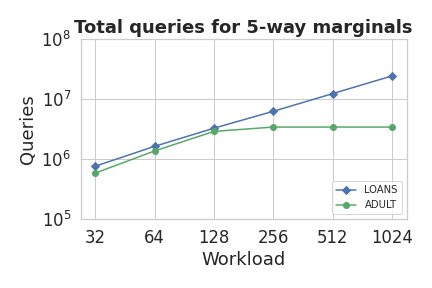}
    \end{tabular}
    \vspace{-.5cm}
    \caption{Total number of queries consistent with the selected random 3-way and 5-way marginals on ADULT and LOANS datasets. Y-axis in log scale.}
    \label{fig:queries}
\end{figure}
 
 Figure \ref{fig:queries} provides the correspondence between the workload size and the number of marginal queries preserved in our experiments. Note that LOANS is a higher dimensional dataset, and so the number of queries continues to increase with the workload, whereas for large enough workloads, we saturate all available queries on ADULT. 

\begin{figure}[t]
    \begin{tabular}{p{0.5\columnwidth}p{0.5\columnwidth}}
        \includegraphics[width=.5\columnwidth,trim={0.5cm 0.5cm 0.5cm 0.5cm}]{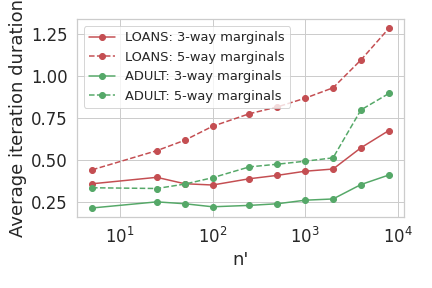} &   \includegraphics[width=.5\columnwidth,trim={0.5cm 0.5cm 0.5cm 0.5cm}]{marginal_figs/stacked_log_scale_running_time.png} \\
        \textbf{(a)} Per-iteration run-time as a function of $n'$ & \textbf{(b)}Total run-time as a function of $n'$ \\[6pt]
        \includegraphics[width=.5\columnwidth,trim={0.5cm 0.5cm 0.5cm 0.5cm}]{marginal_figs/stacked_log_scale.png} & \\
         &  \\
        \textbf{(c)} Error as a function of $n'$ &  \\[6pt]
    \end{tabular}
    \vspace{-.5cm}
    \caption{Run time and error as a function of the synthetic dataset size $n'$. At $n' = 1000$, both total run-time and overall error are near optimal across all settings.}
    \label{fig:runtime}
\end{figure}

Figure \ref{fig:runtime} documents our investigation of the run-time and accuracy of our algorithm as a function of the synthetic dataset size $n'$. $n'$ is a hyperparameter that we can use to trade of the representation ability of our synthetic data (larger $n'$ allows the synthetic data to represent richer sets of answer vectors) with optimization cost. In Figure \ref{fig:runtime} we plot a) the run-time per iteration, b) the total run-time (over all iterations), and c) the error on several datasets and workloads, all as a function of $n'$. We find that although (as expected) the run-time per iteration is monotonically increasing in $n'$, the overall run-time is not --- it grows for sufficiently large $n'$, but also grows for $n'$ that is very small. This seems to be because as our optimization problem becomes sufficiently under-parameterized, the optimization becomes more difficult, and thus our algorithm needs to run for more iterations before convergence. We find that $n' = 1000$ is generally a good choice across datasets and query workloads, simultaneously achieving near minimal error and run-time. Hence we use $n' = 1000$ for all of our other experiments. 

\section{Linear Threshold Functions}

In the body of the paper, we focused on \emph{marginal} queries because of their centrality in the differential privacy literature. But our techniques easily extend to other classes of statistical queries --- all that is required is that we can write python code to evaluate (a differentiable surrogate for) queries in our class. Here we do this for a natural class of linear threshold functions: $t$-out-of-$k$ threshold functions.

\begin{definition}
A $t$-out-of-$k$ threshold query is defined by a subset $S \subseteq [d]$ of $|S| = k$ features, a particular value for each of the features $y \in \prod_{i \in S} \cX_i$, and a threshold $t \leq k$. Given such a pair $(S, y,t)$, the corresponding statistical query $q_{S,y,t}$ is defined as:
$$q_{S,y,t}(x) = \ind(\sum_{i \in S} \ind(x_i = y_i) \geq t)$$
Observe that for each collection of features $S$, there are $\prod_{i \in S}|\cX_i|$ many $t$-out-of-$k$ threshold queries for each threshold $t$. 
\end{definition}

In words, a $t$-out-of-$k$ threshold query evaluates to 1 exactly when at least $t$ of the $k$ features indexed by $S$ take the values indicated by $y$. These generalize the marginal queries that we studied in the body of the paper: A marginal query is simply the special case of a $t$-out-of-$k$ threshold query for $t = k$.

To use our approach to generate synthetic data for $t$-out-of-$k$ linear threshold functions, we need an extended differentiable query class for them. It will be convenient to work with the same one-hot-encoding function $h:\cX\rightarrow \{0,1\}^{d'}$ from the body of the paper, that maps $d$-dimensional vectors of categorical features to $d'$-dimensional vectors of binary features. Our statistical queries are then binary functions defined on the hypercube. We can generically find a differentiable surrogate for our query class by polynomial interpolation: in fact for every boolean function that depends on $k$ variables, there always exists a polynomial of degree $k$ that matches the function on boolean variables, but also extends it in a differentiable manner to the reals. $t$-out-of-$k$ threshold functions are such a class, and so can always be represented by polynomials of degree $k$.

\begin{lemma}
Any boolean class of queries that depends on at most $k$ variables (i.e. a `$k$-junta') has an equivalent extended differentiable query that is a polynomial of degree $k$.
\end{lemma}

In our experiments we will consider $1$-out-of-$k$ queries (equivalently, disjunctions), which have an especially simple extended differentiable representation.

\begin{definition}
Given a subset of features $T \subseteq [d']$, the $1$-out-of-$k$ polynomial query $q_T:\cX^r\rightarrow \mathbb{R}$ is defined as:
$q_T(x) = 1 - \prod_{i \in T} (1-x_i)$.
\end{definition}
It is easy to see that $1$-out-of-$k$ polynomials are extended differentiable queries equivalent to $1$-out-of-$k$ threshold queries. They are differentiable because they are polynomials. A $1$-out-of-$k$ threshold query corresponding to a set of $k$ binary features $T$ (i.e. the one-hot encoded indices for the categorical feature values $y_i$) evaluates to $0$ exactly when \emph{every} binary feature $x_i \in T$ takes value $x_i = 0$ --- i.e. exactly when $\prod_{i \in T}(1-x_1) = 1$. Our $1$-out-of-$k$ polynomials are the negation of this monomial on binary valued inputs. 

The code to evaluate such queries is similarly easy to write --- see Figure \ref{fig:python_snippet2}.

\begin{figure}
\begin{lstlisting}[language=Python, linewidth=\columnwidth,breaklines=true, basicstyle=\small]
import jax.numpy as np
def threeway_thresholded_marginals(D):
  return (D.shape[0] - np.einsum('ij,ik,il->jkl', 1-D, 1-D, 1-D))/D.shape[0]

\end{lstlisting}
\caption{Python function used to compute (an extended equivalent differentiable query for) 1-out-of-3 linear threshold functions}
\label{fig:python_snippet2}
\end{figure}
\begin{figure}[h!]
\center
\includegraphics{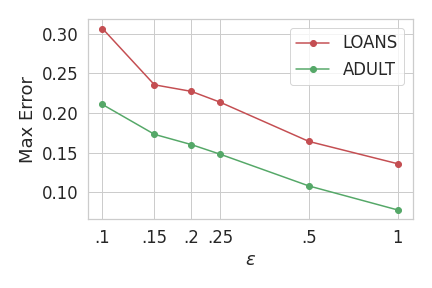}
\caption{Max error for increasing $\epsilon$ of 1-out-of-3 threshold queries with workload 64}
\label{fig:threshold_queries}
\end{figure}

 We repeat our experiments on the Adult and Loans datasets using $1$-out-of-$3$ threshold queries in place of $3$-way marginals. All other experimental details remain the same. In Figure \ref{fig:threshold_queries}, we report the results on a workload of size 64, with $\delta$ fixed to $1/n^2$, and $\epsilon$ ranging from 0.1 to 1.0. 

\end{document}


\twocolumn[
\icmltitle{Supplementary Material for Differentially Private Query Release Through Adaptive Projection}



\icmlsetsymbol{equal}{*}

\begin{icmlauthorlist}
\icmlauthor{Sergul Aydore}{amz}
\icmlauthor{William Brown}{amz,cu}
\icmlauthor{Michael Kearns}{amz,up}
\icmlauthor{Krishnaram Kenthapadi}{amz}
\icmlauthor{Luca Melis}{amz}
\icmlauthor{Aaron Roth}{amz,up}
\icmlauthor{Ankit Siva}{amz}
\end{icmlauthorlist}

\icmlaffiliation{amz}{Amazon AWS AI/ML}
\icmlaffiliation{up}{University of Pennsylvania, Philadelphia, PA, USA}
\icmlaffiliation{cu}{Columbia University, New York, NY, USA}
\icmlcorrespondingauthor{William Brown}{w.brown@columbia.edu}
\icmlcorrespondingauthor{Ankit Siva}{ankitsiv@amazon.com}


\icmlkeywords{Differential Privacy, Synthetic Data}
\vskip 0.3in
]

\renewcommand\thefigure{A.\arabic{figure}}  
\renewcommand\thetable{A.\arabic{table}}   
\renewcommand\thesection{A.\arabic{section}}   
\raggedbottom

\section{Proof of Theorem \ref{thm:accuracy_theorem}}
\label{sec:proof_accuracy_theorem}
\begin{proof}
We reduce to the (unrelaxed) projection mechanism, which has the following guarantee proven by \cite{projection}: for any dataset $D$ consisting of $n$ elements from a \emph{finite} data universe $\cX$, and for any set of $m$ statistical queries $q$, the projection mechanism results in a dataset $D'$ such that: $\sqrt{\frac{1}{m}||q(D')-q(D)||_2^2} \leq \alpha$ for $$\alpha = O\left(\frac{(\ln(|\cX|/\beta)\ln(1/\delta))^{1/4}}{\sqrt{\epsilon n}}\right).$$
Consider a  finite data universe $\cX^\eta = \{0, \eta, 2\eta, \ldots, 1\}^{d'}$ for some discretization parameter $0 < \eta < 1/k$. Given a dataset $D' \in \cX^r$, let $D'_\eta \in \cX^{\eta}$ be the dataset that results from ``snapping'' each real-valued $x \in D$ to its closest discrete valued point $x_\eta \in \cX^r$. Observe that by construction, $||x-x(\eta)||_\infty \leq \eta$, and as a result, for $k$-way product query $q_i$, we have $|q_i(D') - q_i(D'_\eta)| \leq O(\eta k)$. Now let $\hat D' = \arg\min_{\hat D' \in (\cX^r)^*}||a - q(\hat D')||$ and $D'' = \arg\min_{D'' \in (\cX^\eta)^*}||a - q(D'')||$. From above, we know that $\sqrt{\frac{1}{m}||q(D'') - q(\hat D')||} \leq O(\eta k)$, and hence from an application of the triangle inequality, we have that $\sqrt{\frac{1}{m}||q(D) - q(\hat D')||_2^2} \leq O\left(\frac{(\ln(|\cX^\eta|/\beta)\ln(1/\delta))^{1/4}}{\sqrt{\epsilon n}} + \eta k\right)$. Finally, for any dataset $\hat D' \in (\cX^r)^*$, there exists a dataset $D' \in (\cX^r)^{n'}$ such that $\sqrt{\frac{1}{m}||q(D') - q(\hat D')||_2^2} \leq O(\frac{\sqrt{\log k}}{\sqrt{n'}})$ (This follows from a sampling argument, and is proven formally in  \cite{BLR08}.) Hence, a final application of the triangle inequality yields:
$$\sqrt{\frac{1}{m}||q(D) - q(D')||_2^2} \leq$$$$  O\left(\frac{(\ln(|\cX^\eta|/\beta)\ln(1/\delta))^{1/4}}{\sqrt{\epsilon n}} + \eta k + \frac{\sqrt{\log k}}{\sqrt{n'}} \right)$$

Choosing $\eta = \frac{\sqrt{\log k}}{k \sqrt{n'}}$ and noting that $|\cX^\eta| = (\frac{1}{\eta})^{d'}$ yields the bound in our theorem.

\end{proof}

\section{Proof of Theorem \ref{thm:privacy}}
\label{sec:proof_privacy}
\begin{proof}
The privacy of Algorithm \ref{alg:main} follows straightforwardly from the tools we introduced in Section \ref{sec:prelims}. First consider the case of $T = 1$. The algorithm makes $m$ calls to the Gaussian mechanism, each of each satisfies $\rho/m$-zCDP by construction and Lemma \ref{lem:gaussian}. In combination, this satisfies $\rho$-zCDP by the composition Lemma (Lemma \ref{lem:composition}). It then makes a call to the relaxed projection algorithm $RP$, which is a postprocessing of the Gaussian mechanism, and hence does not increase the zCDP parameter, by Lemma \ref{lem:post}. Hence the algorithm is $\rho$-zCDP, and by our choice of $\rho$ and Lemma \ref{lem:conversion}, satisfies $(\epsilon,\delta)$ differential privacy. 

Now consider the case of $T > 1$. Each iteration of the inner loop makes one call to report noisy max, and one call to the Gaussian mechanism. By construction and by Lemmas \ref{lem:gaussian} and \ref{lem:RNM}, each of these calls satisfies $\frac{\rho}{2TK}$-zCDP, and together by the composition Lemma \ref{lem:composition}, satisfy $\frac{\rho}{TK}$-zCDP. The algorithm then makes a call to the relaxed projection algorithm $RP$, which is a post-processing of the composition of the Gaussian mechanism with report noisy max, and so does not increase the zCDP parameter by Lemma \ref{lem:post}. The inner loop runs $T\cdot K$ times, and so the entire algorithm satisfies $\rho$-zCDP by the composition Lemma \ref{lem:composition}. By our choice of $\rho$ and Lemma \ref{lem:conversion}, our algorithm satisfies $(\epsilon,\delta)$ differential privacy as desired. 
\end{proof}

\section{Additional Plots}
\label{sec:extraplots}
\begin{figure}[t]
    \begin{tabular}{cc}
    \includegraphics[width=.48\columnwidth]{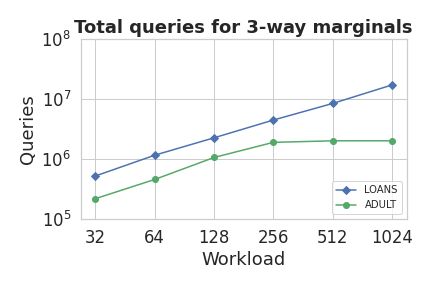} & \includegraphics[width=.48\columnwidth]{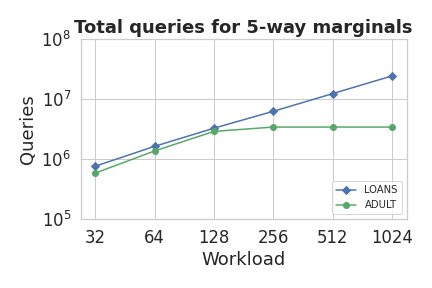}
    \end{tabular}
    \vspace{-.5cm}
    \caption{Total number of queries consistent with the selected random 3-way and 5-way marginals on ADULT and LOANS datasets. Y-axis in log scale.}
    \label{fig:queries}
\end{figure}
 
 Figure \ref{fig:queries} provides the correspondence between the workload size and the number of marginal queries preserved in our experiments. Note that LOANS is a higher dimensional dataset, and so the number of queries continues to increase with the workload, whereas for large enough workloads, we saturate all available queries on ADULT. 

Figure \ref{fig:runtime} docuemnts our investigation of the run-time and accuracy of our algorithm as a function of the synthetic dataset size $n'$. $n'$ is a hyperparameter that we can use to trade of the representation ability of our synthetic data (larger $n'$ allows the synthetic data to represent richer sets of answer vectors) with optimization cost. In Figure \ref{fig:runtime} we plot a) the run-time per iteration, b) the total run-time (over all iterations), and c) the error on several datasets and workloads, all as a function of $n'$. We find that although (as expected) the run-time per iteration is monotonically increasing in $n'$, the overall run-time is not --- it grows for sufficiently large $n'$, but also grows for $n'$ that is very small. This seems to be because as our optimization problem becomes sufficiently under-parameterized, the optimization becomes more difficult, and thus our algorithm needs to run for more iterations before convergence. We find that $n' = 1000$ is generally a good choice across datasets and query workloads, simultaneously achieving near minimal error and run-time. Hence we use $n' = 1000$ for all of our other experiments. 

\begin{figure}[t]
    \begin{tabular}{p{0.5\columnwidth}p{0.5\columnwidth}}
        \includegraphics[width=.5\columnwidth,trim={0.5cm 0.5cm 0.5cm 0.5cm}]{marginal_figs/average_running_time.png} &   \includegraphics[width=.5\columnwidth,trim={0.5cm 0.5cm 0.5cm 0.5cm}]{marginal_figs/stacked_log_scale_running_time.png} \\
        \textbf{(a)} Per-iteration run-time as a function of $n'$ & \textbf{(b)}Total run-time as a function of $n'$ \\[6pt]
        \includegraphics[width=.5\columnwidth,trim={0.5cm 0.5cm 0.5cm 0.5cm}]{marginal_figs/stacked_log_scale.png} & \\
         &  \\ 
        \textbf{(c)} Error as a function of $n'$ &  \\[6pt]
    \end{tabular}
    \vspace{-.5cm}
    \caption{Run time and error as a function of the synthetic dataset size $n'$. At $n' = 1000$, both total run-time and overall error are near optimal across all settings.}
    \label{fig:runtime}
\end{figure}

\section{Linear Threshold Functions}

In the body of the paper, we focused on \emph{marginal} queries because of their centrality in the differential privacy literature. But our techniques easily extend to other classes of statistical queries --- all that is required is that we can write python code to evaluate (a differentiable surrogate for) queries in our class. Here we do this for a natural class of linear threshold functions: $t$-out-of-$k$ threshold functions.

\begin{definition}
A $t$-out-of-$k$ threshold queries is defined by a subset $S \subseteq [d]$ of $|S| = k$ features, a particular value for each of the features $y \in \prod_{i \in S} \cX_i$, and a threshold $t \leq k$. Given such a pair $(S, y,t)$, the corresponding statistical query $q_{S,y,t}$ is defined as:
$$q_{S,y,t}(x) = \ind(\sum_{i \in S} \ind(x_i = y_i) \geq t)$$
Observe that for each collection of features $S$, there are $\prod_{i \in S}|\cX_i|$ many $t$-out-of-$k$ threshold queries for each threshold $t$. 
\end{definition}

In words, a $t$-out-of-$k$ threshold query evaluates to 1 exactly when at least $t$ of the $k$ features indexed by $S$ take the values indicated by $y$. These generalize the marginal queries that we studied in the body of the paper: A marginal query is simply the special case of a $t$-out-of-$k$ threshold query for $t = k$.

To use our approach to generate synthetic data for $t$-out-of-$k$ linear threshold functions, we need an extended differentiable query class for them. It will be convenient to work with the same one-hot-encoding function $h:\cX\rightarrow \{0,1\}^{d'}$ from the body of the paper, that maps $d$-dimensional vectors of categorical features to $d'$-dimensional vectors of binary features. Our statistical queries are then binary functions defined on the hypercube. We can generically find a differentiable surrogate for our query class by polynomial interpolation: in fact for every boolean function that depends on $k$ variables, there always exists a polynomial of degree $k$ that matches the function on boolean variables, but also extends it in a differentiable manner to the reals. $t$-out-of-$k$ threshold functions are such a class, and so can always be represented by polynomials of degree $k$.

\begin{lemma}
Any boolean class of queries that depends on at most $k$ variables (i.e. a `$k$-junta') has an equivalent extended differentiable query that is a polynomial of degree $k$.
\end{lemma}

In our experiments we will consider $1$-out-of-$k$ queries, which have an especially simple extended differentiable representation.

\begin{definition}
Given a subset of features $T \subseteq [d']$, the $1$-out-of-$k$ polynomial query $q_T:\cX^r\rightarrow \mathbb{R}$ is defined as:
$q_T(x) = 1 - \prod_{i \in T} (1-x_i)$.
\end{definition}
It is easy to see that $1$-out-of-$k$ polynomials are extended differentiable queries equivalent to $1$-out-of-$k$ threshold queries. They are differentiable because they are polynomials. A $1$-out-of-$k$ threshold query corresponding to a set of $k$ binary features $T$ (i.e. the one-hot encoded indices for the categorical feature values $y_i$) evaluates to $0$ exactly when \emph{every} binary feature $x_i \in T$ takes value $x_i = 0$ --- i.e. exactly when $\prod_{i \in T}(1-x_1) = 1$. Our $1$-out-of-$k$ polynomials are the negation of this monomial on binary valued inputs. 

The code to evaluate such queries is similarly easy to write --- see Figure \ref{fig:python_snippet2}.

\begin{figure}
\begin{lstlisting}[language=Python, linewidth=\columnwidth,breaklines=true, basicstyle=\small]
import jax.numpy as np
def threeway_thresholded_marginals(D):
  return (D.shape[0] - np.einsum('ij,ik,il->jkl', 1-D, 1-D, 1-D))/D.shape[0]

\end{lstlisting}
\caption{Python function used to compute (an extended equivalent differentiable query for) 1-out-of-3 linear threshold functions}
\label{fig:python_snippet2}
\end{figure}
\begin{figure}[h!]
\center
\includegraphics[width=\columnwidth]{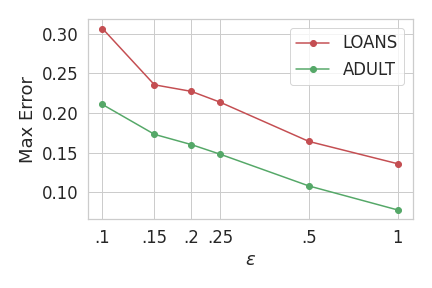}
\caption{Max error for increasing $\epsilon$ of 1-out-of-3 threshold queries with workload 64}
\label{fig:threshold_queries}
\end{figure}

 We repeat our experiments on the Adult and Loans datasets using $1$-out-of-$3$ threshold queries in place of $3$-way marginals. All other experimental details remain the same. In Figure \ref{fig:threshold_queries}, we report the results on a workload of size 64, with $\delta$ fixed to $1/n^2$, and $\epsilon$ ranging from 0.1 to 1.0. 
 
 \section{Ablation studies}
\begin{figure}[t]
        \begin{tabular}{cc}
        {\includegraphics[width=.98\columnwidth]{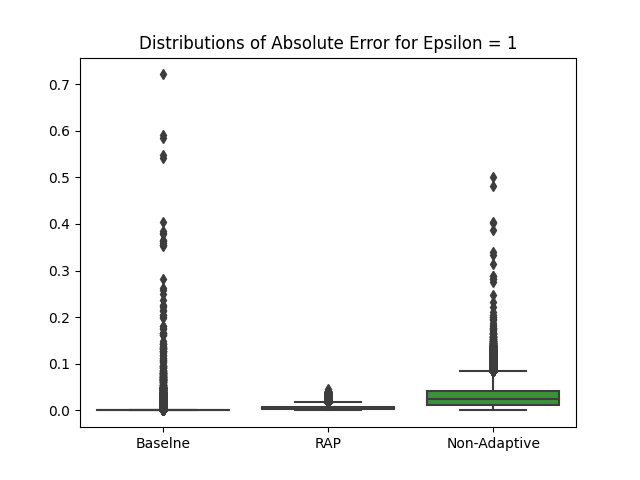}}
      \end{tabular}
    \caption{Box plots of query error distributions on Adult 64 3-way marginals for the nonadaptive ($T=1$) version of our algorithm.}
    \label{fig:epsilons}
\end{figure}

 Our paper contains two innovations on top of the projection mechanism of Nikolov et al. The 1st is to turn the projection into a differentiable optimization problem, and the 2nd is to perform this projection iteratively, restricted to a small number of the highest error remaining queries. The 1st idea is required to make the algorithm implementable, but is not enough to make it competitive with prior work --- See Figure \ref{fig:epsilons}(d) in which we plot the error distribution of the nonadaptive $(T=1)$ version of our algorithm. 

\bibliography{ref}
\bibliographystyle{icml2020}